\documentclass[conference]{IEEEtran}
\IEEEoverridecommandlockouts
\usepackage{cite}
\usepackage{amsmath,amssymb,amsfonts}
\usepackage[noend]{algpseudocode}
\usepackage{algorithmicx}
\usepackage{algorithm}
\usepackage{graphicx}
\usepackage{textcomp}
\usepackage{comment}
\usepackage{float} 

\usepackage{tikz}
\usetikzlibrary{arrows.meta,positioning,shapes.geometric,calc,fit, shadows, backgrounds}

\tikzstyle{block} = [rounded rectangle, minimum width=1cm, minimum height=1cm, text centered, draw=black, thin, inner sep=2pt]
\tikzstyle{arrow} = [thick,->,>=stealth]

\usepackage{color}

\usepackage[utf8]{inputenc} 
\usepackage[T1]{fontenc}    
\usepackage{hyperref}       
\usepackage{url}            
\usepackage{booktabs}       
\usepackage{amsfonts}       
\usepackage{nicefrac}       
\usepackage{microtype}      
\usepackage{xcolor}         
\usepackage{enumitem}
\usepackage{comment}
\usepackage{tcolorbox}
\usepackage{listings}
\usepackage{pdfpages}
\usepackage{longtable}
\usepackage{multirow}
\usepackage{graphicx}
\usepackage{verbatim}
\usepackage{stfloats}
\usepackage[export]{adjustbox}

\usepackage{tabularx}
\usepackage{pifont} 
\usepackage{lipsum} 
\usepackage{colortbl}
\usepackage{float}
\usepackage{hyphenat}

\usepackage{booktabs}
\usepackage{array}
\usepackage{wrapfig}
\usepackage{fancyvrb}

\usepackage{setspace}
\definecolor{swecream}{HTML}{EFFEFF}
\definecolor{issueborder}{HTML}{15071A}
\definecolor{issuefill}{HTML}{F6F8FA}
\definecolor{envfill}{HTML}{F2F9FF}
\definecolor{envborder}{HTML}{123C7C}
\definecolor{agentfill}{HTML}{F9F3F3}
\definecolor{agentborder}{HTML}{9B0A0A}
\definecolor{goldpatchborder}{HTML}{FABB00}
\definecolor{goldpatchfill}{HTML}{FFF7E1}

\DefineVerbatimEnvironment{CodeVerbatim}{Verbatim}{
  formatcom={\color{black}},
  fontsize=\small,
  fontfamily=\ttdefault,
  fontseries=\mddefault,
  fontshape=\updefault,
  fillcolor=\color{white},
  framerule=0pt,
}

\newtcolorbox{observationbox}[1][]{
        colback=envfill,
        colbacktitle=envfill,
        colframe=envborder,
        arc=5pt,
        fontupper=\small,
        fonttitle=\bfseries\color{black},
        boxrule=0.5mm,
        boxsep=1mm,
        width=\linewidth,
        breakable,
        title={Observation \hfill #1},
        rounded corners,
        toptitle=0.7mm,
        bottomtitle=0.7mm
}
\newtcolorbox{goldpatchbox}[1][]{
        colback=goldpatchfill,
        colbacktitle=goldpatchfill,
        colframe=goldpatchborder,
        arc=5pt,
        fontupper=\small,
        fonttitle=\bfseries\color{black},
        boxrule=0.5mm,
        boxsep=1mm,
        width=\linewidth,
        breakable,
        title={\twemoji{1f6a9} Flag Captured \hfill #1},
        rounded corners,
        toptitle=0.7mm,
        bottomtitle=0.7mm
}
\newtcolorbox{issuebox}[1][]{
        colback=issuefill,
        colbacktitle=issuefill,
        colframe=issueborder,
        arc=5pt,
        fontupper=\small,
        fonttitle=\bfseries\color{black},
        boxrule=0.5mm,
        boxsep=1mm,
        width=\linewidth,
        breakable,
        title={Issue \hfill #1},
        rounded corners,
        toptitle=1mm
}
\newtcolorbox{agentbox}[1][]{
        colback=agentfill,
        colbacktitle=agentfill,
        colframe=agentborder,
        arc=5pt,
        fontupper=\small,
        fonttitle=\bfseries\color{black},
        boxrule=0.5mm,
        boxsep=1mm,
        width=\linewidth,
        breakable,
        title={EnIGMA \hfill #1},
        rounded corners,
        toptitle=1mm,
        lower separated=false
}
\newtcolorbox{fileviewerbox}[1]{
        enhanced,
        breakable,
        boxrule = 1.5pt,
        fontupper = \small,
        fonttitle = \bf\color{black},
        arc = 5pt,
        rounded corners,
        colframe = black,
        colbacktitle = swecream,
        colback = swecream,
        title = #1,
        left=4pt 
}
\newtcolorbox{promptbox}[1]{
    enhanced,
    breakable,
    boxrule=1pt,  
    fontupper=\small,
    fonttitle=\bfseries\color{black},
    arc=3pt,  
    rounded corners,
    colframe=black,
    colbacktitle=swecream,
    colback=swecream,
    title=#1,
    left=2mm,  
    right=2mm,  
    top=1mm,  
    bottom=1mm  
}

\def\BibTeX{{\rm B\kern-.05em{\sc i\kern-.025em b}\kern-.08em
    T\kern-.1667em\lower.7ex\hbox{E}\kern-.125emX}}
\begin{document}

\title{MVB-Grasp: Minimum-Volume-Box Filtering of Diffusion-based Grasps for Frontal Manipulation}


\author{
\IEEEauthorblockN{Bibek Poudel\textsuperscript{*}, Abdul Basit\textsuperscript{*}, Muhammad Shafique}
\IEEEauthorblockA{\textit{eBRAIN Lab, Division of Engineering} \textit{New York University (NYU) Abu Dhabi}, Abu Dhabi, UAE\\
\{bp2376, abdul.basit, muhammad.shafique\}@nyu.edu}
\thanks{* Authors have equal contributions.}
\vspace{-10pt}
}

\maketitle

\begin{abstract}

State-of-the-art 6-DoF grasp generators excel on tabletop benchmarks with overhead cameras but struggle in frontal grasping scenarios on low-cost manipulators with constrained workspaces, where kinematic limits and approach-direction constraints cause high failure rates.
We address this challenge for the Unitree Z1 arm by proposing \textbf{MVB-Grasp}, a novel grasping stack that injects a Minimum Volume Bounding Box (MVBB) geometric prior into diffusion-based grasp generation to dramatically improve success rates in frontal, workspace-constrained settings.
Our key scientific contributions are threefold: 
(i) an MVBB-based geometric filter that exploits oriented bounding-box face normals to reject grasps approaching through the table or misaligned with accessible object faces in $\mathcal{O}(N)$ time; 
(ii) a combined re-scoring function that blends learned discriminator scores with face-alignment geometry ($\alpha=0.85$), specifically calibrated for the Z1's frontal workspace and kinematic constraints; and 
(iii) a systematic MuJoCo evaluation protocol measuring grasp success across object types, distances, lateral positions, and pitch orientations to validate embodiment-specific performance.
We implement MVB-Grasp on a Unitree Z1 arm with an Intel RealSense D405 camera, integrating YOLOv8 object detection, GraspGen for candidate generation, Principal Component Analysis (PCA)-based MVBB fitting, and inverse-kinematics trajectory planning.
Experiments across 81 MuJoCo episodes (cylinder, asymmetric box, waterbottle) demonstrate that MVB-Grasp achieves 59.3\% success versus 24.7\% for vanilla GraspGen, a \textbf{2.4$\times$ improvement}, by filtering geometrically infeasible candidates and prioritizing face-aligned grasps suited to the Z1's frontal approach constraints. Real-world trials confirm that the MVBB prior substantially improves grasp reliability on constrained, low-cost manipulators without requiring model retraining.
\end{abstract}


\section{Introduction}

Grasping is the primary interface between a robot and its environment: without reliable grasp acquisition, manipulation skills such as rearrangement, tool use, and pick--place are infeasible in everyday settings~\cite{newbury2022review}. Recent advances in 6-DoF grasp synthesis have produced diverse deep models operating on point clouds and RGB--D data, including sampling-based methods (GPD, Contact-GraspNet)~\cite{mahler2017gpd,contactgraspnet}, dense predictors (AnyGrasp)~\cite{anygrasp}, and diffusion-based generators (GraspGen)~\cite{graspgen}. These systems achieve strong performance on tabletop and cluttered-scene benchmarks, typically under top-down or mildly oblique viewpoints.

However, deploying such powerful 6-DoF grasp generators on low-cost manipulators with constrained workspaces remains challenging. Most state-of-the-art methods are trained and evaluated in scenarios with overhead cameras and gravity-aligned approaches, but affordable arms such as the Unitree Z1 often operate in \emph{frontal grasping} settings with side-mounted cameras and tight kinematic constraints. In these scenarios, many generated grasps are geometrically infeasible, approaching through the support surface, deeply penetrating object faces, or requiring joint configurations beyond the arm's limits. The embodiment-agnostic nature of general-purpose generators means they cannot anticipate which approach directions are accessible for a specific manipulator, leading to high failure rates despite strong learned discriminators.

\begin{figure}[t]
    \centering
    \includegraphics[width=\linewidth]{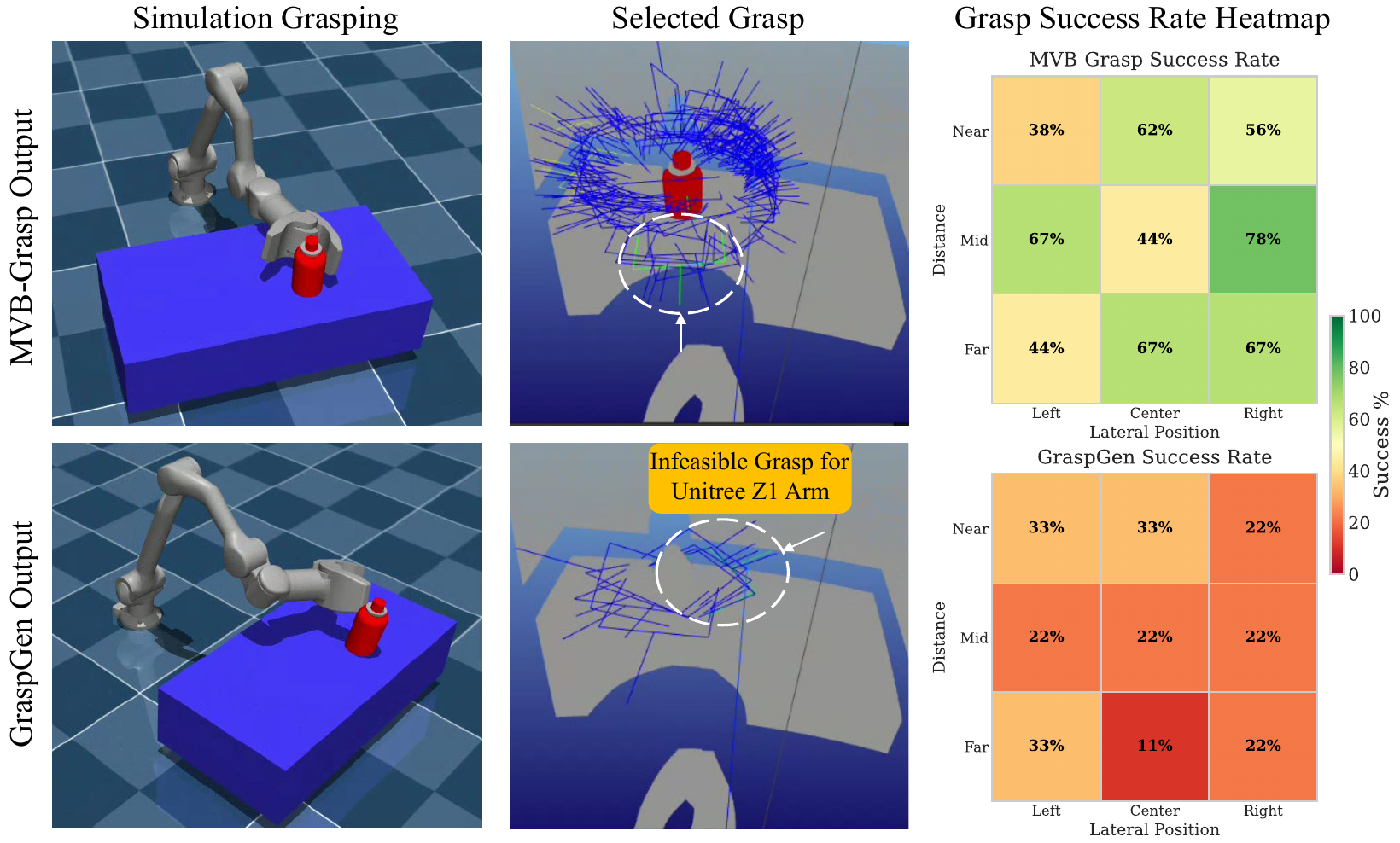}
    \caption{Motivation for MVB-Grasp. 
    Top: the proposed MVB-Grasp selects stable frontal grasps that succeed in simulation and achieve consistently high success rates across the workspace. 
    Bottom: the vanilla GraspGen baseline often selects less robust grasps, yielding lower and spatially inconsistent success. 
    From left to right we show the simulated grasp execution, the distribution of sampled and selected grasps, and the empirical success-rate heatmaps over distance--lateral bins for each method.}
    \label{fig:motivation}
\end{figure}

\subsection{Motivational case study: frontal grasping failures on Z1}

We study tabletop frontal grasping with the Unitree Z1 arm using a side-mounted RGB--D camera, integrating GraspGen with both MuJoCo simulation and a calibrated real setup. In its vanilla form, GraspGen samples $K$ grasp candidates, ranks them with a learned discriminator, and applies collision checks before execution.

Two observations motivate this work. First, many Z1 failures, such as through-table approaches, deep penetrations, or joint-infeasible poses, are geometrically evident from the point cloud, yet are ranked highly by GraspGen’s embodiment-agnostic discriminator. Second, simple geometric cues, including object face normals and principal axes, often suffice to distinguish accessible from infeasible approaches in the Z1’s frontal workspace (Fig.~\ref{fig:motivation}).

In this setting, a PCA-fitted oriented bounding box (OBB), which we use as a practical MVBB approximation, captures key object geometry, including extents, support plane, and dominant face normals. Many invalid grasps can be rejected using OBB/MVBB alignment alone, without physics simulation. Thus, improving frontal 6-DoF grasping on Z1 reduces to three key questions: how to design a geometric prior that filters infeasible grasps while preserving valid ones, how to integrate this prior with general-purpose generators like GraspGen without retraining, and how to characterize when such geometry helps, or hurts, grasp success.

\subsection{Scientific challenges targeted in this paper}

Towards this, we formalize three scientific challenges. 
First, \emph{geometric priors for embodiment-specific filtering}: how simple object geometry (e.g., OBB/MVBB) can remove grasps that are infeasible for a frontal, workspace-limited arm while preserving sufficient grasp diversity. 
Second, \emph{training-free embodiment adaptation}: how to re-score candidates from a general grasp generator such as GraspGen so that they favor the Z1’s frontal workspace without retraining the generator or its discriminator. 
Third, \emph{when geometry helps, or hurts}: under what object shapes and viewing conditions geometric priors improve grasp success, and when they over-constrain generation, especially for asymmetric or close-range objects.

\subsection{Our contributions}

We propose a training-free alignment strategy that augments an off-the-shelf 6-DoF grasp generator (GraspGen) for frontal grasping with the Unitree Z1 arm. Our contributions are:
\begin{enumerate}
\item \textbf{OBB-based alignment prior for frontal 6-DoF grasping:} We apply a PCA-based OBB, used as a lightweight MVBB prior, to extract face normals, filter misaligned grasps, and re-score candidates by combining geometric alignment with GraspGen’s discriminator. This training-free prior integrates with a diffusion-based generator and improves frontal grasp success on a workspace-constrained arm.

\item \textbf{Structured frontal-grasp evaluation on Unitree Z1:} We introduce an 81-scenario MuJoCo benchmark and compare vanilla GraspGen with the OBB-augmented pipeline, analyzing success rates, failure modes, and candidate statistics across distance, lateral position, occlusion, and object orientation.

\item \textbf{Geometry--learning trade-off analysis:} We study per-object, distance, and orientation effects, identifying when alignment priors help (e.g., symmetric objects) and when they over-constrain generation (e.g., asymmetric objects at close range).
\end{enumerate}

Overall, injecting this OBB prior into GraspGen yields a 2.4$\times$ improvement in simulated success (59.3\% vs.\ 24.7\%) with only 6.78\,ms additional latency, making the approach practical for real-time deployment. Although our experiments focus on GraspGen and the Z1 arm, the OBB module is architecture-agnostic and provides a template for embodiment-specific tuning of general-purpose grasp perception systems.

\section{Related work and background}

Robotic grasping has evolved from analytic planning to learned vision-based synthesis. Early methods~\cite{LenzGraspIJRR2015,RedmonGraspICRA2015,MorrisonGGCNN2018,MahlerDexNet2RSS2017} predict planar grasp rectangles in image space assuming top-down approaches with overhead cameras. While highly successful for bin-picking and similar setups, these approaches are fundamentally limited to near-planar (4-DoF) grasps and do not directly support arbitrary 6-DoF poses or frontal approaches.

Modern 6-DoF systems address this through three main paradigms.  
(1) \emph{Sampling-based generators} such as 6-DoF GraspNet and Contact-GraspNet sample grasp poses in $SE(3)$ and refine them using learned quality metrics and collision checks~\cite{Mousavian6DoFGraspNetICCV2019,contactgraspnet}. They offer strong success in clutter but are designed for general scenarios rather than specific embodiments.  
(2) \emph{Dense prediction methods} including AnyGrasp predict grasp scores densely over the visible scene in a single feed-forward pass~\cite{anygrasp}, enabling temporally smooth execution and robustness to depth noise, but still requiring downstream collision checking and planning specific to the robot.  
(3) \emph{Generative models} such as diffusion-based GraspGen model the distribution of successful grasps conditioned on point clouds~\cite{graspgen}. These models can generate diverse grasps across gripper types, but are typically deployed as general modules without embodiment-specific tuning or explicit latency-aware design.

Table~\ref{tab:related-compare-grasp} summarizes representative frameworks along two axes that are central to this work: (i) the grasp representation (planar vs.\ 6-DoF), and (ii) the dominant approach mode (top-down vs.\ frontal/general). Most frameworks are optimized for top-down or general 6-DoF scenarios; none explicitly target frontal grasping with a Z1-like arm.

\begin{table}[ht]
  \centering
  \caption{Comparison of representative vision-based grasp frameworks.}
  \label{tab:related-compare-grasp}
  \small
  \resizebox{\columnwidth}{!}{%
    \begin{tabular}{lcc}
      \toprule
      \textbf{Framework} & \textbf{Grasp DoF} & \textbf{Approach mode} \\
      \midrule
      Lenz \emph{et al.}~\cite{LenzGraspIJRR2015} & Planar (4-DoF) & Top-down \\
      Redmon \& Angelova~\cite{RedmonGraspICRA2015} & Planar (4-DoF) & Top-down \\
      GG-CNN~\cite{MorrisonGGCNN2018} & Planar (4-DoF) & Top-down / wrist \\
      Dex-Net~2.0~\cite{MahlerDexNet2RSS2017} & Planar (4-DoF) & Top-down bin picking \\
      6-DoF GraspNet~\cite{Mousavian6DoFGraspNetICCV2019} & 6-DoF samples & General 6-DoF \\
      Contact-GraspNet~\cite{contactgraspnet} & 6-DoF dense & General 6-DoF \\
      AnyGrasp~\cite{anygrasp} & 6/7-DoF dense & Multi-view, mostly top-down \\
      GraspGen~\cite{graspgen} & 6-DoF diffusion & General 6-DoF \\
      MVBB grasping~\cite{HuebnerMVBBICRA2008,GeidenstamBoxBasedRSS2009} & Planar (4-DoF) & Various \\
      GoalGrasp~\cite{ZhangGoalGrasp2024} & 6-DoF & General 6-DoF \\
      \midrule
      \textbf{MVB-Grasp (Ours)} & \textbf{6-DoF + MVBB} & \textbf{Frontal, Z1-specific} \\
      \bottomrule
    \end{tabular}%
  }
\\[0.3em]
{\footnotesize
Grasp DoF: grasp representation (planar vs.\ full 6-DoF). 
Approach mode: dominant camera viewpoint and approach direction in evaluations.
}
\end{table}

A complementary line of work exploits geometric primitives as priors for grasp planning. H\"ubner \emph{et al.}~\cite{HuebnerMVBBICRA2008} decompose objects into MVBBs for planar grasping in simulation, and Geidenstam \emph{et al.}~\cite{GeidenstamBoxBasedRSS2009} learn 2D strategies from box-based approximations. More recent methods use oriented boxes with machine learning~\cite{LiOBBLightGBM2021,VoOBBYOLO2025}, but still target planar grasps. In 6-DoF settings, GoalGrasp~\cite{ZhangGoalGrasp2024} uses 3D box priors for user-specified objects under occlusion, and AnyGrasp or Contact-GraspNet employ geometry mainly for cropping and simple clearance checks~\cite{anygrasp,contactgraspnet}. In these systems, geometric reasoning primarily serves to remove obviously invalid candidates or localize objects; the core scoring models remain generic and are not tuned to a specific arm’s workspace, preferred approach directions (e.g., frontal vs.\ vertical), or latency budget.

Most prior work therefore optimizes grasp \emph{accuracy and generality} across diverse robots and scenarios, while embodiment-specific constraints and end-to-end latency are treated as downstream filtering problems. Recent efforts that encode physical or geometric priors for 6-DoF grasping~\cite{ma2024physicalpriors,heatmap6dof} show that injecting structure can improve robustness, but they do not specifically address the \emph{frontal grasping} challenge on workspace-constrained, low-cost manipulators like the Z1.

\section{Methodology}
\label{sec:method}

This section presents the design and implementation of \textbf{MVB-Grasp}, a latency-aware grasping pipeline that integrates a MVBB prior with a state-of-the-art diffusion-based grasp generator. We begin with a system overview, then describe each component in detail: perception, grasp generation, the MVBB module, grasp filtering and re-scoring, and Z1 arm execution.

\subsection{System Overview}
\label{subsec:system-overview}

Fig.~\ref{fig:methodology} illustrates the complete MVB-Grasp pipeline. Given an RGB-D observation from a RealSense D405 camera mounted on or near the Unitree Z1 arm, our system performs the following stages:

\begin{figure}[ht]
    \centering
    \includegraphics[width=\linewidth]{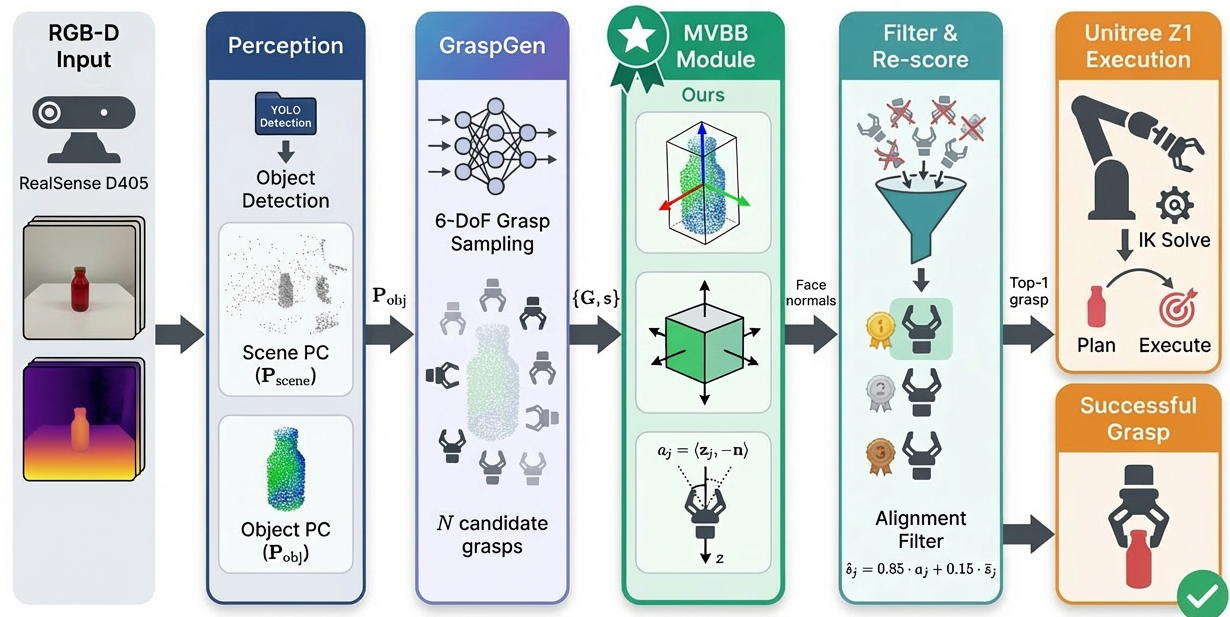}
    \caption{Overview of MVB-Grasp: RGB-D images are processed to obtain an object point cloud, GraspGen samples 6-DoF candidate grasps, our MVBB module extracts face normals to filter and re-score candidates for frontal alignment, and the top-ranked grasp is executed on the Unitree Z1.}
    \label{fig:methodology}
\end{figure}

\begin{enumerate}
    \item \textbf{Perception}: Acquire RGB-D frames, detect the target object via YOLO, and segment the object point cloud $\mathcal{P}_\text{obj}$ from the scene point cloud $\mathcal{P}_\text{scene}$.
    \item \textbf{Grasp Generation}: Feed $\mathcal{P}_\text{obj}$ into GraspGen~\cite{graspgen} to sample $N$ candidate 6-DoF grasp poses $\{(\mathbf{G}_j, s_j)\}_{j=1}^{N}$, where $\mathbf{G}_j \in SE(3)$ and $s_j \in [0,1]$ is the discriminator confidence.
    \item \textbf{MVBB Module}: Fit an OBB to $\mathcal{P}_\text{obj}$, extract face normals, and compute alignment scores between each grasp's approach axis and the nearest OBB faces.
    \item \textbf{Filtering \& Re-scoring}: Reject grasps that do not align with any accessible face; re-score remaining grasps by combining geometric alignment with the original confidence.
    \item \textbf{Execution}: Select the top-ranked grasp, solve inverse kinematics, plan a collision-free trajectory, and execute on the Z1 arm.
\end{enumerate}

The key innovation is that the MVBB-based geometric prior (stages 3--4) can aggressively prune infeasible grasps in $\mathcal{O}(N)$ time, often eliminating the need for expensive physics-based collision checking that scales as $\mathcal{O}(N \cdot |\mathcal{P}_\text{scene}|)$.

\subsection{Perception Pipeline}
\label{subsec:perception}

\subsubsection{RGB-D Acquisition}
We use an Intel RealSense D405 depth camera operating at $640 \times 480$ resolution at 30~FPS. The depth stream is aligned to the color stream to ensure pixel correspondence. Let $I_\text{rgb} \in \mathbb{R}^{H \times W \times 3}$ denote the RGB image and $I_\text{depth} \in \mathbb{R}^{H \times W}$ the aligned depth map (in meters).

\subsubsection{Object Detection}
We employ YOLOv8~\cite{yolov8_ultralytics} for real-time object detection. Given $I_\text{rgb}$, the detector outputs a set of bounding boxes $\mathcal{B} = \{(x_1^{(i)}, y_1^{(i)}, x_2^{(i)}, y_2^{(i)}, c^{(i)})\}$ for objects of interest, where $c^{(i)}$ is the detection confidence. In our experiments, we typically target a single object (e.g., a bottle) and select the highest-confidence detection.

\subsubsection{Point Cloud Generation and Segmentation}
We convert the depth image to a 3D point cloud using the camera intrinsics $(f_x, f_y, c_x, c_y)$:
\begin{equation}
    \mathbf{p}_{u,v} = \begin{bmatrix}
        (u - c_x) \cdot d_{u,v} / f_x \\
        (v - c_y) \cdot d_{u,v} / f_y \\
        d_{u,v}
    \end{bmatrix}, \quad d_{u,v} = I_\text{depth}(u,v)
    \label{eq:depth-to-pc}
\end{equation}
where $(u, v)$ are pixel coordinates. Points with invalid depth ($d_{u,v} \leq 0$ or $d_{u,v} > d_\text{max}$) are discarded.

The scene point cloud is:
$
    \mathcal{P}_\text{scene} = \{\mathbf{p}_{u,v} \mid d_{u,v} \in (0, d_\text{max}]\}
$

The object point cloud is segmented by retaining only points whose pixel coordinates fall within the detected bounding box:
\begin{equation}
    \mathcal{P}_\text{obj} = \{\mathbf{p}_{u,v} \in \mathcal{P}_\text{scene} \mid x_1 \leq u \leq x_2, \; y_1 \leq v \leq y_2\}
\end{equation}

To remove outliers and noise from $\mathcal{P}_\text{obj}$, we apply:
\begin{itemize}
    \item \textbf{Depth percentile clipping}: Remove points outside the $[1\%, 99\%]$ depth percentile to eliminate spurious spikes.
    \item \textbf{Statistical outlier removal}: Remove points whose mean distance to $k$ nearest neighbors exceeds $\mu + 2\sigma$.
    \item \textbf{Radius outlier removal}: Remove points with fewer than $n_\text{min}$ neighbors within radius $r$.
\end{itemize}

\subsection{GraspGen Integration}
\label{subsec:graspgen}

We integrate GraspGen~\cite{graspgen}, a diffusion-based 6-DoF grasp generator that operates on point clouds. GraspGen models the distribution of successful grasps conditioned on the object geometry using a denoising diffusion probabilistic model (DDPM) coupled with a PointTransformerV3 backbone.

\subsubsection{Grasp Representation}
Each grasp pose is represented as a rigid transformation $\mathbf{G} \in SE(3)$, parameterized as a $4 \times 4$ homogeneous matrix:
\begin{equation}
    \mathbf{G} = \begin{bmatrix}
        \mathbf{R} & \mathbf{t} \\
        \mathbf{0}^\top & 1
    \end{bmatrix}, \quad \mathbf{R} \in SO(3), \; \mathbf{t} \in \mathbb{R}^3
\end{equation}
where $\mathbf{R}$ is the grasp orientation and $\mathbf{t}$ is the grasp position. The grasp frame convention follows the Robotiq 2F-140 gripper: the $z$-axis points along the approach direction (into the palm), the $y$-axis is parallel to the finger opening direction, and the $x$-axis completes the right-handed frame.

\subsubsection{Sampling and Scoring}
Given $\mathcal{P}_\text{obj}$, we invoke GraspGen to sample $N$ candidate grasps:
$\{(\mathbf{G}_j, s_j)\}_{j=1}^{N} = \textsc{GraspGen}(\mathcal{P}_\text{obj}, N)$ where $s_j \in [0,1]$ is the discriminator score indicating grasp quality. In our experiments, we use $N = 800$ grasps per scene. The scores are normalized to $[0,1]$:
\begin{equation}
    \bar{s}_j = \frac{s_j - s_\text{min}}{s_\text{max} - s_\text{min} + \epsilon}
    \label{eq:score-normalize}
\end{equation}
where $s_\text{min} = \min_j s_j$, $s_\text{max} = \max_j s_j$, and $\epsilon = 10^{-6}$ prevents division by zero.

\subsection{MVBB Module}
\label{subsec:mvbb}

The core contribution of MVB-Grasp is the MVBB module, which exploits the geometry of an oriented bounding box fitted to the object point cloud to filter and re-score grasp candidates.

\subsubsection{Oriented Bounding Box Fitting}
Given $\mathcal{P}_\text{obj} = \{\mathbf{p}_i\}_{i=1}^{M}$, we compute the MVBB using PCA followed by convex hull optimization.

First, we compute the centroid:
$
    \boldsymbol{\mu} = \frac{1}{M} \sum_{i=1}^{M} \mathbf{p}_i
$

The covariance matrix is:
\begin{equation}
    \boldsymbol{\Sigma} = \frac{1}{M-1} \sum_{i=1}^{M} (\mathbf{p}_i - \boldsymbol{\mu})(\mathbf{p}_i - \boldsymbol{\mu})^\top
\end{equation}

Eigendecomposition yields the principal axes:
\begin{equation}
    \boldsymbol{\Sigma} = \mathbf{V} \boldsymbol{\Lambda} \mathbf{V}^\top, \quad \mathbf{V} = [\mathbf{v}_1 \mid \mathbf{v}_2 \mid \mathbf{v}_3]
\end{equation}
where $\mathbf{v}_1, \mathbf{v}_2, \mathbf{v}_3$ are the eigenvectors ordered by decreasing eigenvalue $\lambda_1 \geq \lambda_2 \geq \lambda_3$.

The OBB is defined by:
\begin{itemize}
    \item \textbf{Rotation matrix}: $\mathbf{R}_\text{obb} = [\mathbf{v}_1 \mid \mathbf{v}_2 \mid \mathbf{v}_3] \in SO(3)$
    \item \textbf{Center}: $\mathbf{c}_\text{obb} \in \mathbb{R}^3$ (refined from $\boldsymbol{\mu}$)
    \item \textbf{Extents}: $\mathbf{e} = (e_1, e_2, e_3)$ giving half-lengths along each principal axis
\end{itemize}

The transformation from the world frame to the OBB frame:
\begin{equation}
    \mathbf{T}_\text{obb} = \begin{bmatrix}
        \mathbf{R}_\text{obb} & \mathbf{c}_\text{obb} \\
        \mathbf{0}^\top & 1
    \end{bmatrix}
    \label{eq:obb-transform}
\end{equation}

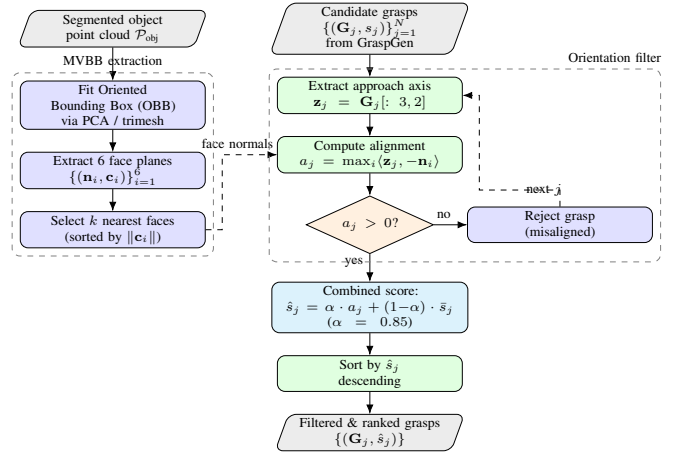
\begin{figure}[ht]
\centering
\resizebox{\columnwidth}{!}{
\begin{tikzpicture}[
    node distance = {4mm and 8mm},
    every node/.style = {font=\scriptsize},
    io/.style = {trapezium, trapezium left angle=70, trapezium right angle=110,
                 draw, fill=gray!15, inner sep=2pt, text width=28mm, rounded corners,
                 align=center},
    procA/.style = {rectangle, draw, fill=blue!12, rounded corners,
                    inner sep=2pt, text width=32mm, align=center},
    procB/.style = {rectangle, draw, fill=green!12, rounded corners,
                    inner sep=2pt, text width=32mm, align=center},
    comb/.style = {rectangle, draw, fill=cyan!12, rounded corners,
                   inner sep=2pt, text width=34mm, align=center},
    dec/.style = {diamond, draw, fill=orange!12, aspect=2.2,
                  inner sep=0.8pt, text width=16mm, align=center},
    group/.style = {draw, rounded corners, inner sep=4pt, dashed, gray},
    ->, >=Latex
]

\node[io] (objpc) {Segmented object\\point cloud $\mathcal{P}_\text{obj}$};

\node[procA, below=6mm of objpc] (obb) {Fit Oriented\\Bounding Box (OBB)\\via PCA / trimesh};

\node[procA, below=of obb] (extract) {Extract 6 face planes\\$\{(\mathbf{n}_i, \mathbf{c}_i)\}_{i=1}^{6}$};

\node[procA, below=of extract] (select) {Select $k$ nearest faces\\(sorted by $\|\mathbf{c}_i\|$)};

\node[io, right=14mm of objpc] (grasps) {Candidate grasps\\$\{(\mathbf{G}_j, s_j)\}_{j=1}^{N}$\\from GraspGen};

\node[procB, below=of grasps] (zaxis) {Extract approach axis\\$\mathbf{z}_j = \mathbf{G}_j[:3,2]$};

\node[procB, below=of zaxis] (align) {Compute alignment\\$a_j = \max_i \langle \mathbf{z}_j, -\mathbf{n}_i \rangle$};

\node[dec, below=of align] (thresh) {$a_j > 0$?};

\node[comb, below=5mm of thresh] (score) {Combined score:\\$\hat{s}_j = \alpha \cdot a_j + (1{-}\alpha) \cdot \bar{s}_j$\\($\alpha = 0.85$)};

\node[procB, below=of score] (rank) {Sort by $\hat{s}_j$\\descending};

\node[io, below=of rank] (out) {Filtered \& ranked grasps\\$\{(\mathbf{G}_j, \hat{s}_j)\}$};

\node[procA, right=6mm of thresh] (reject) {Reject grasp\\(misaligned)};

\draw (objpc) edge (obb)
      (obb) edge (extract)
      (extract) edge (select);

\draw (grasps) edge (zaxis)
      (zaxis) edge (align)
      (align) edge (thresh);

\draw (thresh) -- node[left, pos=0.3]{\scriptsize yes} (score);
\draw (thresh) -- node[above, pos=0.3]{\scriptsize no} (reject);

\coordinate (rejectUp) at ($(reject.north)+(0,3mm)$);
\coordinate (loopBack) at ($(zaxis.east)+(3mm,0)$);
\draw[->, dashed] (reject.north) -- (rejectUp) -| node[right, pos=0.25]{\scriptsize next $j$} (loopBack) -- (zaxis.east);

\draw (score) edge (rank)
      (rank) edge (out);

\coordinate (selectRight) at ($(select.east)+(3mm,0)$);
\coordinate (alignLeft) at ($(align.west)+(-3mm,0)$);
\draw[->, dashed] (select.east) -- (selectRight) |- node[above, pos=0.7]{\scriptsize face normals} (alignLeft) -- (align.west);

\node[group, fit=(obb) (select), label={[font=\scriptsize]above:MVBB extraction}] {};
\node[group, fit=(zaxis) (thresh) (reject), label={[font=\scriptsize]above right:Orientation filter}] {};

\end{tikzpicture}
}
\caption{MVBB-based grasp filtering and scoring. Left: MVBB extraction from the object point cloud. Right: Each candidate grasp is checked for alignment with the nearest OBB faces; aligned grasps receive a combined score mixing geometric alignment ($\alpha=0.85$) and the original GraspGen confidence.}
\label{fig:mvbb-filtering}
\end{figure}

\subsubsection{Face Extraction}
The OBB has six faces, each defined by a face normal $\mathbf{n}$ and face center $\mathbf{c}$. For each principal axis $i \in \{1, 2, 3\}$ and sign $\sigma \in \{-1, +1\}$:
\begin{align}
    \mathbf{n}_{i,\sigma} &= \sigma \cdot \mathbf{v}_i \label{eq:face-normal} \\
    \mathbf{c}_{i,\sigma} &= \mathbf{c}_\text{obb} + \frac{e_i}{2} \cdot \mathbf{n}_{i,\sigma} \label{eq:face-center}
\end{align}

This yields six face descriptors:
\begin{equation}
    \mathcal{F}_\text{all} = \{(\mathbf{n}_{i,\sigma}, \mathbf{c}_{i,\sigma}) \mid i \in \{1,2,3\}, \sigma \in \{-1,+1\}\}
\end{equation}

While the above equations define the MVBB module analytically, Fig.~\ref{fig:mvbb-filtering} provides a procedural view of how it operates. Starting from the segmented object point cloud, the left column shows OBB fitting and extraction of the six face planes, followed by the selection of the $k$ faces closest to the camera. The right column illustrates how, for each GraspGen candidate, we compute the approach axis, evaluate its alignment to the selected faces, reject misaligned grasps, and assign the combined score $\hat{s}_j$ that blends geometric alignment with the normalized GraspGen confidence. This flow directly corresponds to the MVBB-related calls in Algorithm~\ref{alg:mvbgrasp-pipeline}.

\subsubsection{Face Selection}
Not all faces are accessible for frontal grasping on a tabletop. We select the $k$ nearest faces (typically $k=2$) by sorting face centers by their distance to the camera origin (or world origin): $\mathcal{F} = \textsc{TopK}\left(\mathcal{F}_\text{all}, k, \text{key} = \|\mathbf{c}_{i,\sigma}\|\right)$

This heuristic prioritizes faces that are closer to the robot (and thus more likely to be reachable), while also implicitly avoiding the bottom face resting on the table.

\subsection{Grasp Filtering and Re-scoring}
\label{subsec:filtering}

The MVBB module enables two key operations: (1) filtering out geometrically infeasible grasps, and (2) re-scoring surviving grasps to favor those aligned with accessible object faces.

\subsubsection{Alignment Computation}
For each grasp $\mathbf{G}_j$, we extract the approach axis (the $z$-column of the rotation matrix): $\mathbf{z}_j = \mathbf{G}_j[:3, 2] = \mathbf{R}_j \mathbf{e}_z
    \label{eq:approach-axis}$ where $\mathbf{e}_z = [0, 0, 1]^\top$ is the unit vector along the local $z$-axis.

A grasp approaching a face should have its approach axis pointing \emph{into} the face, i.e., opposite to the outward face normal. The alignment score with face $(\mathbf{n}, \mathbf{c})$ is: $a(\mathbf{z}_j, \mathbf{n}) = \langle \mathbf{z}_j, -\mathbf{n} \rangle = -\mathbf{z}_j^\top \mathbf{n}$. The best alignment over all selected faces is:
\begin{equation}
    a_j = \max_{(\mathbf{n}, \mathbf{c}) \in \mathcal{F}} a(\mathbf{z}_j, \mathbf{n}) = \max_{(\mathbf{n}, \mathbf{c}) \in \mathcal{F}} \left( -\mathbf{z}_j^\top \mathbf{n} \right)
    \label{eq:alignment-best}
\end{equation}

Note that $a_j \in [-1, 1]$, where $a_j = 1$ indicates perfect alignment (grasp approaches perpendicular to a face), and $a_j \leq 0$ indicates the grasp is pointing away from all selected faces.

\subsubsection{Hard Filtering}
We reject grasps that do not align with any accessible face: $\mathcal{G}_\text{pass} = \{(\mathbf{G}_j, s_j) \mid a_j > 0\}$. This simple threshold eliminates grasps that approach through the table, from behind the object, or at extreme tangential angles. In practice, this filter typically removes 15--35\% of candidates, depending on object shape and pose.

\subsubsection{Combined Re-scoring}
For grasps that pass the hard filter, we compute a combined score that blends the geometric alignment with the original GraspGen confidence: $\hat{s}_j = \alpha \cdot a_j + (1 - \alpha) \cdot \bar{s}_j$
where $\alpha \in [0, 1]$ controls the trade-off between geometric alignment and learned quality. We use $\alpha = 0.85$ in all experiments, emphasizing geometric feasibility for the Z1 arm's constrained workspace.

\subsubsection{Ranking}
The filtered grasps are ranked by descending combined score: $\mathcal{G}_\text{ranked} = \textsc{SortDescending}(\mathcal{G}_\text{pass}, \text{key} = \hat{s})$. The top-$k$ grasps (typically $k=1$ or $k=3$) are passed to the execution stage.

\subsection{Z1 Arm Execution}
\label{subsec:execution}

\subsubsection{Coordinate Frame Transformations}
The grasp poses from GraspGen are expressed in the camera frame $\{C\}$. To execute on the Z1 arm, we transform to the robot base frame $\{B\}$: $\mathbf{G}_j^{B} = \mathbf{T}_{BC} \cdot \mathbf{G}_j^{C}$, where $\mathbf{T}_{BC} \in SE(3)$ is the camera-to-base extrinsic calibration.

\subsubsection{Inverse Kinematics}
The Z1 arm has 6 revolute joints. Given a target end-effector pose $\mathbf{G}^{B}_\text{target}$, we solve for joint angles $\mathbf{q} \in \mathbb{R}^6$ using iterative inverse kinematics:
\begin{equation}
    \mathbf{q}^* = \arg\min_{\mathbf{q}} \| \textsc{FK}(\mathbf{q}) - \mathbf{G}^{B}_\text{target} \|_{SE(3)}^2 + \lambda \|\mathbf{q} - \mathbf{q}_\text{home}\|^2
    \label{eq:ik}
\end{equation}
subject to joint limits $\mathbf{q}_\text{min} \leq \mathbf{q} \leq \mathbf{q}_\text{max}$, where $\textsc{FK}(\cdot)$ is the forward kinematics and $\lambda$ is a regularization term favoring solutions near the home configuration.

If IK fails for the top-ranked grasp, we proceed to the next candidate in $\mathcal{G}_\text{ranked}$.

\subsubsection{Trajectory Planning and Execution}
Once a valid IK solution is found, we plan a trajectory from the current configuration $\mathbf{q}_\text{curr}$ to the pre-grasp pose (offset along the approach axis), then to the final grasp pose:
\begin{enumerate}
    \item Move to pre-grasp: $\mathbf{G}_\text{pre} = \mathbf{G}^{B}_\text{target} \cdot \mathbf{T}_\text{offset}$, where $\mathbf{T}_\text{offset}$ translates $-d_\text{pre}$ along the $z$-axis.
    \item Approach: Linear motion from $\mathbf{G}_\text{pre}$ to $\mathbf{G}^{B}_\text{target}$.
    \item Close gripper.
    \item Lift: Retract along the $z$-axis by $d_\text{lift}$.
\end{enumerate}

\subsubsection{Optional Collision Checking}
For deployments where additional safety is required, we optionally perform collision checking between the gripper mesh and $\mathcal{P}_\text{scene}$. For each grasp candidate $\mathbf{G}_j$, we transform the gripper collision mesh vertices $\mathcal{V}_\text{gripper}$ to the world frame and check for penetration:
\begin{equation}
    \text{collision}(\mathbf{G}_j) = \exists \mathbf{v} \in \mathcal{V}_\text{gripper}^{\mathbf{G}_j} : \min_{\mathbf{p} \in \mathcal{P}_\text{scene}} \|\mathbf{v} - \mathbf{p}\| < \tau_\text{coll}
    \label{eq:collision-check}
\end{equation}
where $\tau_\text{coll}$ is the collision threshold (typically 2~mm). However, a key hypothesis of MVB-Grasp is that the MVBB filtering makes collision checking optional for many scenarios, substantially reducing latency.

\subsection{Complexity Analysis}
\label{subsec:complexity}

The computational complexity of each pipeline stage depends on the input size:
\begin{itemize}
    \item \textbf{Perception}: Depth-to-pointcloud conversion and YOLO detection scale as $\mathcal{O}(HW)$ with image resolution.
    \item \textbf{GraspGen inference}: Scales as $\mathcal{O}(M \cdot N)$ where $M$ is the object point cloud size and $N$ is the number of generated grasps.
    \item \textbf{OBB fitting}: PCA-based fitting is $\mathcal{O}(M)$ linear in point cloud size.
    \item \textbf{Alignment filtering}: Checking $N$ grasps against $k$ faces is $\mathcal{O}(N \cdot k)$, which is very fast since $k \ll N$.
    \item \textbf{Collision checking}: The bottleneck in vanilla pipelines, scaling as $\mathcal{O}(N \cdot |\mathcal{V}| \cdot |\mathcal{P}_\text{scene}|)$ where $|\mathcal{V}|$ is the number of gripper mesh vertices.
\end{itemize}

The key insight is that MVBB-based filtering ($\mathcal{O}(N \cdot k)$) is orders of magnitude faster than physics-based collision checking. Detailed latency measurements are presented in Section~\ref{sec:results}.

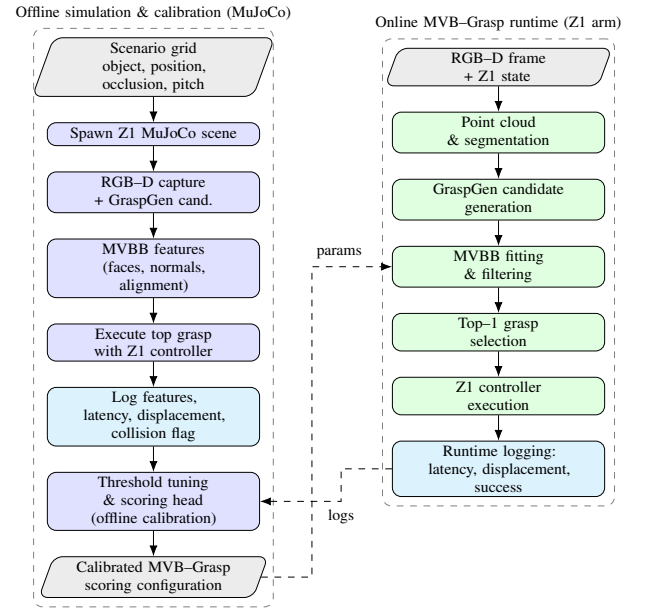
\begin{figure}[ht]
\centering
\resizebox{0.95\columnwidth}{!}{
\begin{tikzpicture}[
    node distance = {4mm and 10mm},
    every node/.style = {font=\scriptsize},
    io/.style = {trapezium, trapezium left angle=70, trapezium right angle=110,
                 draw, fill=gray!15, inner sep=2pt, text width=30mm,
                 rounded corners, align=center},
    procA/.style = {rectangle, draw, fill=blue!12, rounded corners,
                    inner sep=2pt, text width=32mm, align=center},
    procB/.style = {rectangle, draw, fill=green!12, rounded corners,
                    inner sep=2pt, text width=32mm, align=center},
    comb/.style  = {rectangle, draw, fill=cyan!12, rounded corners,
                    inner sep=2pt, text width=32mm, align=center},
    dec/.style   = {diamond, draw, fill=orange!12, aspect=2,
                    inner sep=0.8pt, text width=14mm, align=center},
    group/.style = {draw, rounded corners, inner sep=4pt, dashed, gray},
    ->, >=Latex
]

\node[io] (grid) {Scenario grid\\object, position,\\occlusion, pitch};

\node[procA,below=of grid] (spawn) {Spawn Z1 MuJoCo scene};

\node[procA,below=of spawn] (simgen) {RGB--D capture\\+ GraspGen cand.};

\node[procA,below=of simgen] (simmvbb) {MVBB features\\(faces, normals,\\alignment)};

\node[procA,below=of simmvbb] (simexec) {Execute top grasp\\with Z1 controller};

\node[comb,below=of simexec] (log) {Log features,\\latency, displacement,\\collision flag};

\node[procA,below=of log] (tune) {Threshold tuning\\\& scoring head\\(offline calibration)};

\node[io,below=of tune] (cfg) {Calibrated MVB--Grasp\\scoring configuration};

\node[group,fit=(grid) (cfg),label=above:{\scriptsize Offline simulation \& calibration (MuJoCo)}] {};

\node[io,right=20mm of grid] (rgbd) {RGB--D frame\\+ Z1 state};

\node[procB,below=of rgbd] (pc) {Point cloud\\\& segmentation};

\node[procB,below=of pc] (gen) {GraspGen candidate\\generation};

\node[procB,below=of gen] (mvbb) {MVBB fitting\\\& filtering};

\node[procB,below=of mvbb] (sel) {Top--1 grasp\\selection};

\node[procB,below=of sel] (exec) {Z1 controller\\execution};

\node[comb,below=of exec] (rtlog) {Runtime logging:\\latency, displacement,\\success};

\node[group,fit=(rgbd) (rtlog),label=above:{\scriptsize Online MVB--Grasp runtime (Z1 arm)}] {};

\draw (grid)  -- (spawn);
\draw (spawn) -- (simgen);
\draw (simgen) -- (simmvbb);
\draw (simmvbb) -- (simexec);
\draw (simexec) -- (log);
\draw (log) -- (tune);
\draw (tune) -- (cfg);

\draw (rgbd) -- (pc);
\draw (pc)   -- (gen);
\draw (gen)  -- (mvbb);
\draw (mvbb) -- (sel);
\draw (sel)  -- (exec);
\draw (exec) -- (rtlog);

\coordinate (midCfg) at ($(cfg.east)+(8mm,0)$);
\coordinate (midMvbb) at ($(mvbb.west)+(-4mm,0)$);
\draw[dashed] (cfg.east) -- (midCfg) |- node[above,pos=0.75]{\scriptsize params} (midMvbb) -- (mvbb.west);

\coordinate (midLog) at ($(rtlog.west)+(-8mm,0)$);
\coordinate (midTune) at ($(tune.east)+(4mm,0)$);
\draw[dashed] (rtlog.west) -- (midLog) |- node[below,pos=0.5]{\scriptsize logs} (midTune) -- (tune.east);

\end{tikzpicture}
}
\caption{Offline simulation and calibration pipeline (left) and online MVB--Grasp runtime on the Z1 arm (right). Solid arrows indicate forward execution; dashed arrows indicate feedback from runtime logs and calibrated parameters.}
\label{fig:mvbgrasp-flow}
\end{figure}

The same MVB-Grasp stack is used both for systematic evaluation in MuJoCo and for online execution on the physical Z1. 
Fig.~\ref{fig:mvbgrasp-flow} visualizes this dual use: the left branch shows the offline simulation and calibration loop, and the right branch shows the online runtime pipeline deployed on the robot.

\begin{algorithm}[ht]
\scriptsize
\caption{MVB-Grasp: Latency-aware 6-DoF grasping with MVBB prior}
\label{alg:mvbgrasp-pipeline}
\begin{algorithmic}[1]
\Require $\mathcal{C}, \mathcal{G}, \mathcal{A}, \mathcal{D}, \mathcal{H}, \alpha, k, N$
\Ensure $\mathbf{G}^*$

\State $(I_{\text{rgb}}, I_{\text{depth}}) \gets \mathcal{C}.\textsc{Capture}()$
\State $\mathcal{P}_{\text{obj}} \gets \textsc{SegmentObject}(\textsc{DepthToPointCloud}(I_{\text{depth}}), \mathcal{D}.\textsc{Detect}(I_{\text{rgb}}))$

\State $\{(\mathbf{G}_j,s_j)\}_{j=1}^{N} \gets \mathcal{G}.\textsc{Sample}(\mathcal{P}_\text{obj},N)$
\State $\bar{s}_j \gets \textsc{Normalize}(\{s_j\})$

\State $\mathcal{F} \gets \textsc{FitMVBBAndNearestFaces}(\mathcal{P}_\text{obj},k)$

\State $\mathcal{G}_\text{filtered} \gets \{(\mathbf{G}_j,\alpha a_j + (1-\alpha)\bar{s}_j)\;|\; a_j>0\}$
\Statex \hspace{1em} where $a_j=\max_{(\mathbf{n},\cdot)\in\mathcal{F}} \langle \mathbf{G}_j[:3,2],-\mathbf{n} \rangle$

\State $\mathcal{G}_\text{ranked} \gets \textsc{SortDescending}(\mathcal{G}_\text{filtered})$

\For{$(\mathbf{G},\hat{s}) \in \mathcal{G}_\text{ranked}$}
    \State $\mathbf{q} \gets \textsc{IKSolve}(\mathcal{A},\mathbf{G})$
    \If{$\mathbf{q}\neq\texttt{None}$ \textbf{and} 
        (\textbf{not} \textsc{CollisionCheckEnabled} \textbf{or} 
        \textbf{not} \textsc{CheckCollision}$(\mathcal{H},\mathbf{G},\mathcal{P}_\text{scene})$)}
        \State $\mathbf{G}^*\gets\mathbf{G}$; \textbf{break}
    \EndIf
\EndFor
\State \textsc{PlanAndExecute}$(\mathcal{A},\mathbf{q},\mathbf{G}^*)$
\State \Return $\mathbf{G}^*$
\end{algorithmic}
\end{algorithm}

\section{Experimental Setup}
\subsection{Simulation Setup (MuJoCo)}
\label{sec:setup_sim}

\begin{figure}[ht]
  \centering
  \begin{tabular}{@{}cc@{}}
    \includegraphics[height=3.2cm,keepaspectratio]{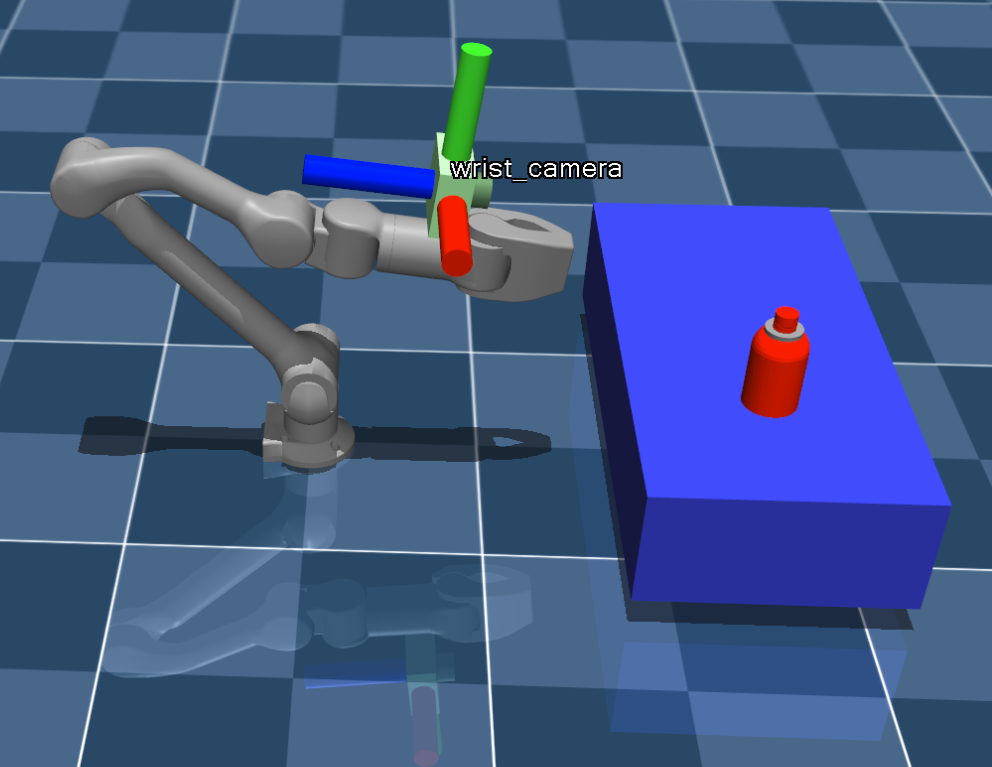} &
    \includegraphics[height=3.2cm,keepaspectratio]{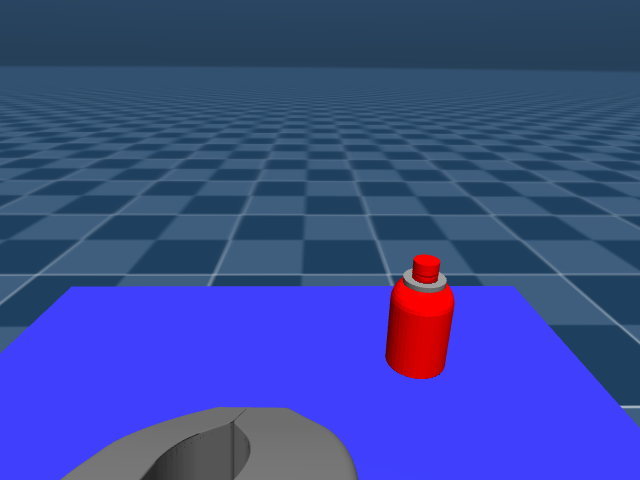}
  \end{tabular}
  \caption{MuJoCo simulation setup: (a) Unitree Z1 Pro arm in a tabletop scene, (b) wrist-mounted RGB--D view used for point cloud reconstruction.}
  \label{fig:sim_mujoco}
\end{figure}

\textbf{Environment and scenario grid:}
All simulation experiments are run in MuJoCo with a physics-accurate Unitree Z1 Pro and
parallel-jaw gripper in a tabletop workspace (Fig.~\ref{fig:sim_mujoco}).
Each episode contains a single object (cylinder, asymmetric box, or water bottle)
placed on the table according to a structured grid with:
\emph{distance} (Near, Mid, Far),
\emph{lateral offset} (Left, Center, Right),
and \emph{pitch} ($-45^\circ$, $0^\circ$, $+45^\circ$),
yielding 81 distinct configurations detailed in Algorithm~\ref{alg:sim-evaluation}.
\\

\begin{algorithm}[ht]
\scriptsize
\caption{Systematic grasp evaluation using MuJoCo}
\label{alg:sim-evaluation}
\begin{algorithmic}[1]
\Require Objects $\mathcal{O}$; grid $\mathcal{X}=\{\text{Near, Mid, Far}\}\times\{\text{Left, Center, Right}\}$;
         occlusion levels $\mathcal{L}$; pitches $\Theta$;
         MuJoCo scene $\mathcal{S}$; camera pose $\mathbf{T}_{\rm cam}$;
         GraspGen model $\mathcal{G}$; methods $\mathcal{M}=\{\textsc{Vanilla},\textsc{MVB-Grasp}\}$
\Ensure Result log $\mathcal{R}$
\State $\mathcal{R}\gets\emptyset$
\For{$o\in\mathcal{O}$,\ $(x,y)\in\mathcal{X}$,\ $\ell\in\mathcal{L}$,\ $\theta\in\Theta$}
  \State \textsc{ResetScene}$(\mathcal{S})$; \ \textsc{SpawnObject}$(\mathcal{S},o,x,y,\theta,\ell)$
  \State $(I_{\rm rgb},I_{\rm depth})\gets\textsc{RenderRGBD}(\mathcal{S},\mathbf{T}_{\rm cam})$
  \State $P_{\rm obj},P_{\rm scene}\gets\textsc{SegmentPointClouds}(I_{\rm depth},o)$
  \State $\{(G_j,s_j)\}_{j=1}^N \gets \mathcal{G}.\textsc{Sample}(P_{\rm obj},N)$
  \For{$m\in\mathcal{M}$}
    \If{$m=\textsc{Vanilla}$}
      \State $G_{\rm ranked}\gets\textsc{SortByScore}(\textsc{CollisionFilter}(\{(G_j,s_j)\},P_{\rm scene}))$
    \Else \Comment{MVB-Grasp}
      \State $F\gets\textsc{ComputeOBBFaces}(P_{\rm obj})$
      \State $G_{\rm ranked}\gets\textsc{MVBBFilterAndRescore}(\{(G_j,s_j)\},F)$
    \EndIf
    \State \textbf{success} $\gets$ \textsc{ExecuteTopK}$(\mathcal{S},G_{\rm ranked},k,\tau_{\rm lift})$
    \State Append $(o,x,y,\ell,\theta,m,\textbf{success},|G_{\rm ranked}|)$ to $\mathcal{R}$
  \EndFor
\EndFor
\State \Return $\mathcal{R}$
\end{algorithmic}
\end{algorithm}



\begin{figure}[ht]
  \centering
  \begin{tabular}{@{}cc@{}}
    \includegraphics[height=3.2cm,width=3.2cm]{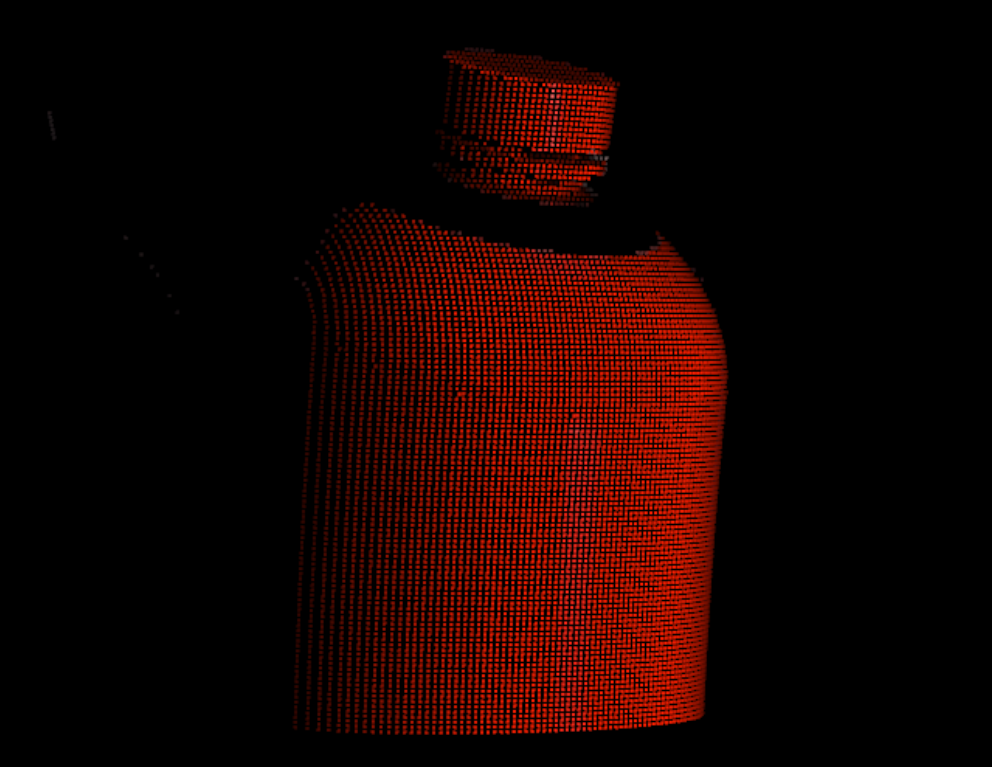} &
    \includegraphics[height=3.2cm,width=3.2cm]{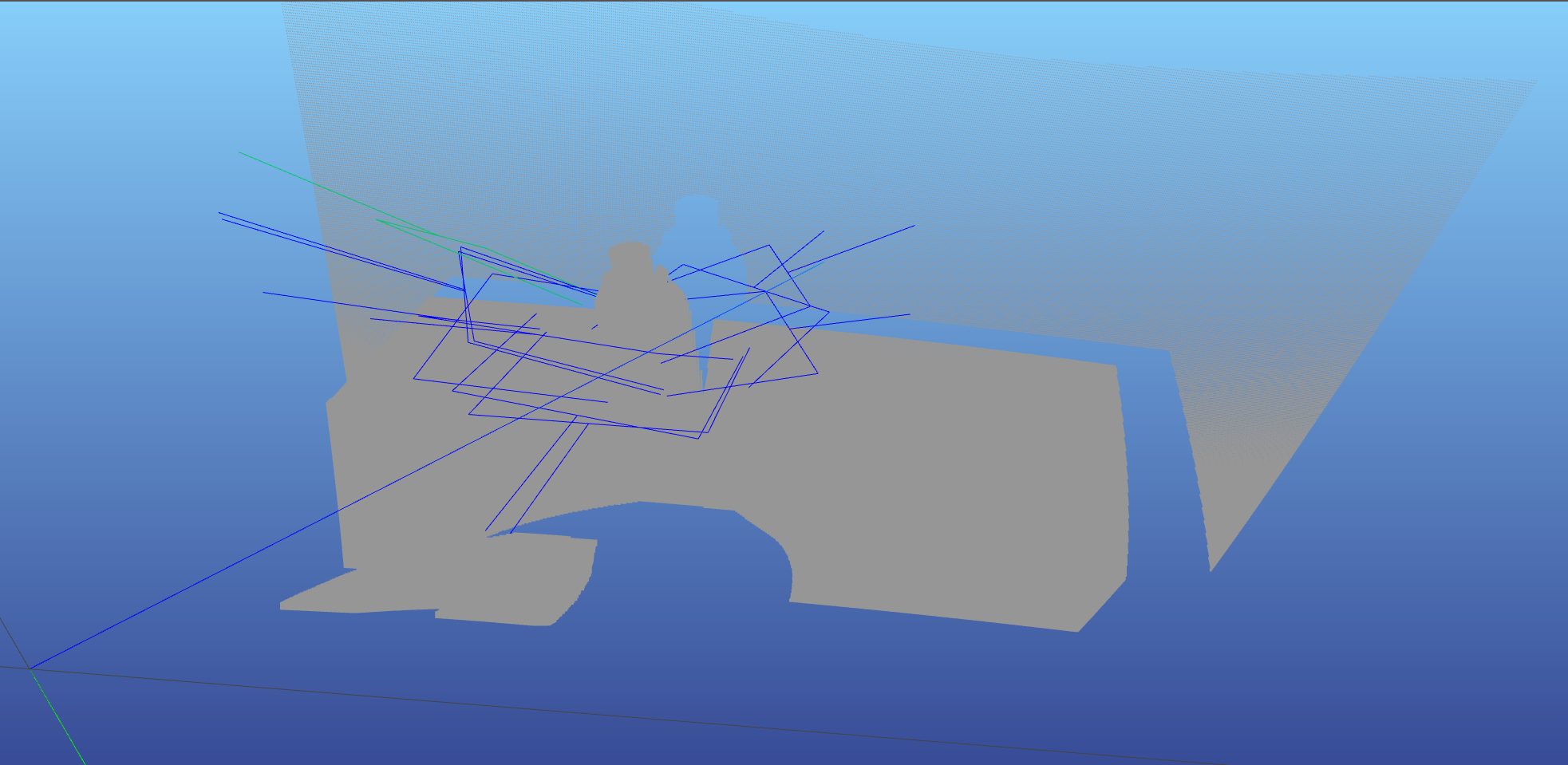}
  \end{tabular}
  \caption{Perception and grasp candidates in simulation:
  (a) segmented object point cloud,
  (b) example GraspGen 6-DoF grasp proposals.}
  \label{fig:sim_pc_grasps}
\end{figure}

\textbf{Perception and candidate generation:}
From a fixed home configuration $q_{\text{home}} = [0, 1.085, -0.261, -0.523, 0, 1.57]$~rad,
we render synchronized RGB--D images, back-project to a scene point cloud,
and segment the target object using the known mask.
The segmented point cloud is passed to GraspGen to generate
$K$ 6-DoF grasp candidates per episode (Fig.~\ref{fig:sim_pc_grasps}).

\begin{figure}[ht]
  \centering
  \includegraphics[height=3.2cm,keepaspectratio]{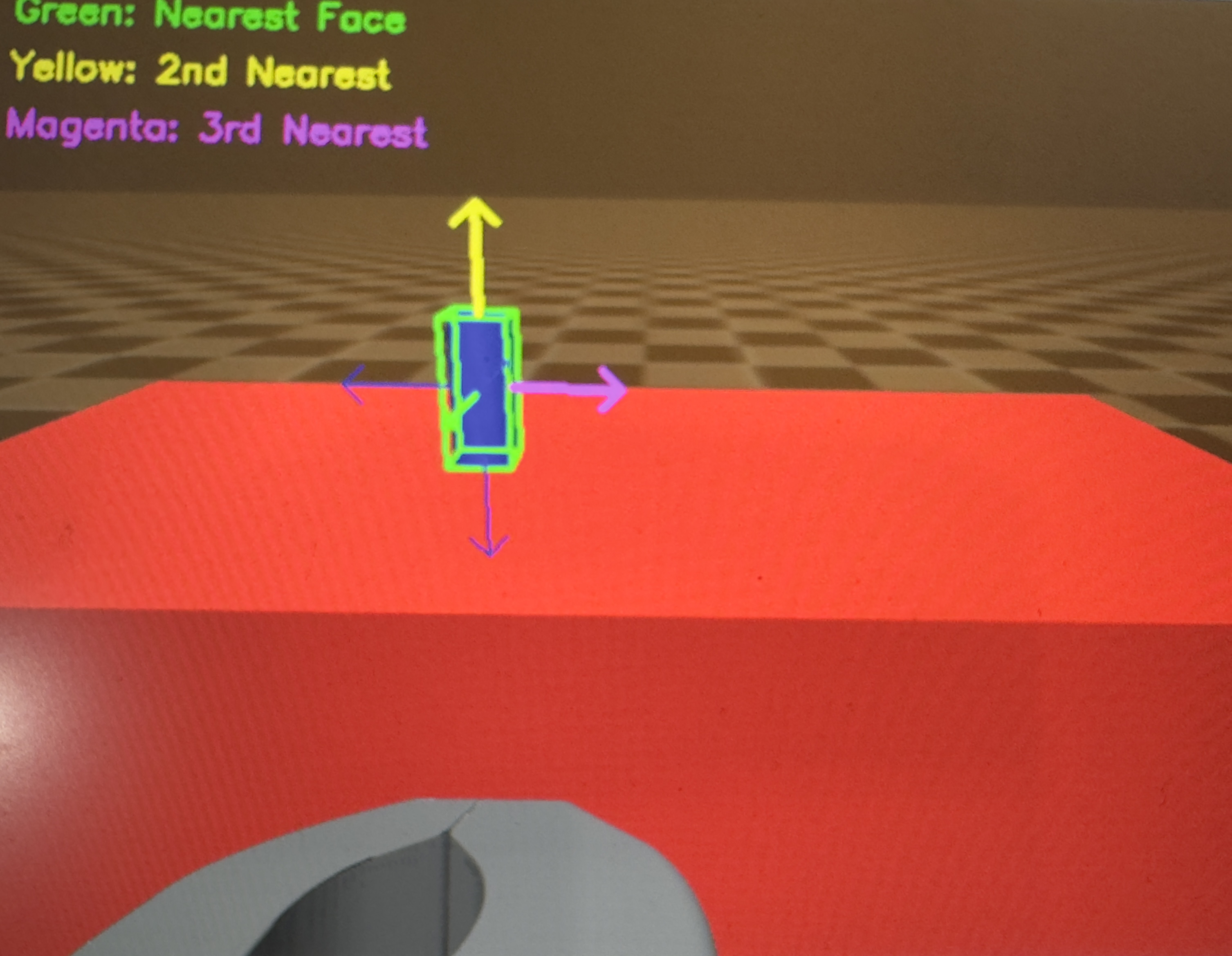}
  \vspace{-5pt}
  \caption{MVBB in simulation: fitted oriented box and three selected nearest faces used for alignment-based filtering. \vspace{-5pt}}
  \label{fig:mvbb_sim}
\end{figure}

\textbf{MVBB filtering and execution:}
An oriented MVBB is fitted to the object point cloud to obtain face centers and normals (Fig.~\ref{fig:mvbb_sim}).  
The $k$ faces closest to the camera are selected, grasp candidates are filtered by approach alignment to these faces, and surviving grasps are re-scored using the MVBB alignment score and GraspGen confidence.
The top-ranked grasp is transformed to the robot base frame and executed via a Cartesian trajectory (LERP position, SLERP orientation).
In simulation, a grasp is counted as successful if the object is lifted and stably retained after the closing and lift motion.

\begin{figure}[ht]
  \centering
  \includegraphics[width=0.5\textwidth]{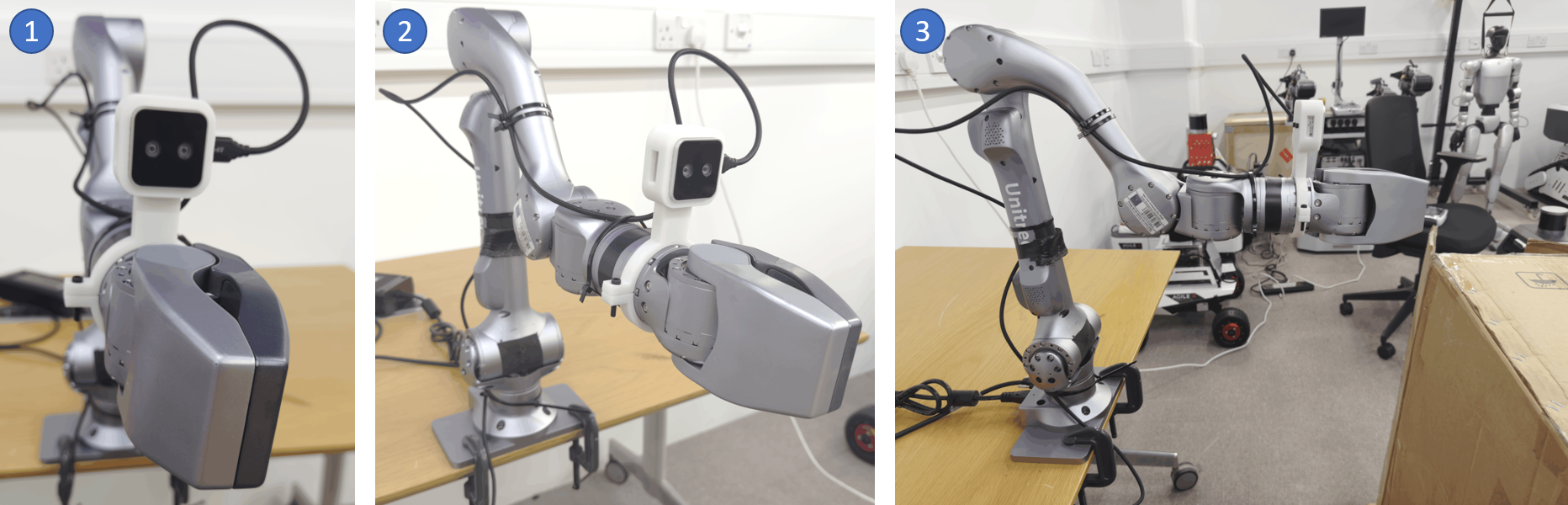}
  \vspace{-5pt}
  \caption{Hardware setup: (1, 2) Unitree Z1 Pro arm with Intel RealSense D405 camera, (3) tabletop view mirroring the MuJoCo configuration. \vspace{-5pt}}
  \label{fig:real_setup}
\end{figure}

\textbf{Platform and calibration:}
Real-world experiments use a Unitree Z1 Pro with a wrist-mounted Intel RealSense D405 RGB--D camera (Fig.~\ref{fig:real_setup}).  
Camera intrinsics and hand--eye calibration are obtained using a ChArUco board, and the table and camera poses are arranged to closely match the MuJoCo setup for sim-to-real comparison.

\subsection{Real-World Setup (Z1 Hardware)}
\label{sec:setup_real}

\begin{figure}[ht]
  \centering
  \begin{minipage}[t]{0.48\linewidth}
    \centering
    \includegraphics[height=3.2cm,width=\linewidth,keepaspectratio]{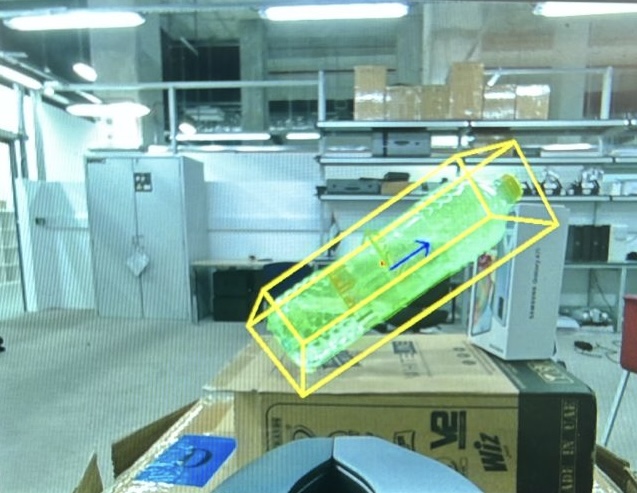}
  \end{minipage}
  \hfill
  \begin{minipage}[t]{0.48\linewidth}
    \centering    \includegraphics[height=3.2cm,width=\linewidth,keepaspectratio]{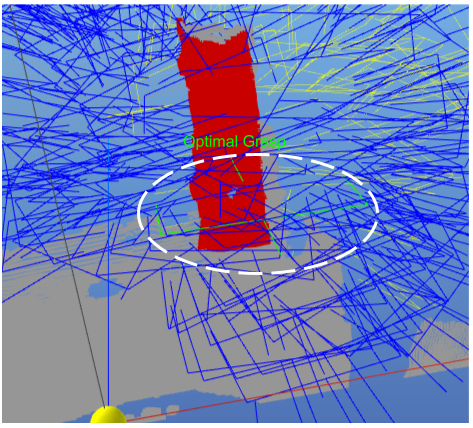}
  \end{minipage}
  \caption{Real-world MVBB pipeline: (a) MVBB fitted to the segmented bottle point cloud, (b) filtered grasp set with selected grasp shown in green.}
  \label{fig:mvbb_real_pipeline}
\end{figure}

\textbf{Object placement and execution.}
We consider a single target object (water bottle) placed within the reachable workspace with pose sampled from the same distance--lateral--pitch grid as in simulation.
RGB--D frames from the calibrated camera are back-projected to a scene point cloud, the bottle is segmented, and MVB-Grasp runs the same GraspGen + MVBB pipeline as in MuJoCo (Fig.~\ref{fig:mvbb_real_pipeline}).
The top-ranked grasp is executed with conservative approach and lift distances to maintain table clearance and avoid joint limits.

\begin{figure}
    \centering
    \includegraphics[width=\linewidth]{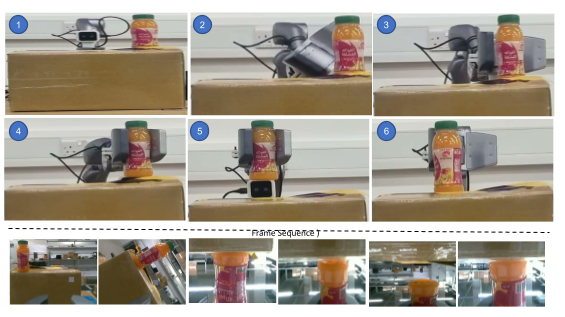}
    \caption{Example real-world run: MVB-Grasp executing a frontal grasp on the Z1 arm.}
    \label{fig:real_run}
\end{figure}

\begin{figure}[h]
  \centering
  \includegraphics[width=\linewidth]{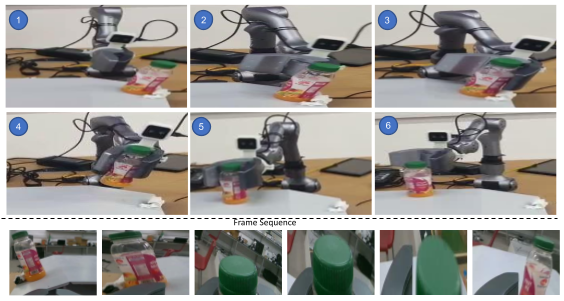}
  \caption{Qualitative slanted-object grasps with MVB-Grasp in the real world.}
  \label{fig:slanted_obj_collage}
\end{figure}

\textbf{Evaluation protocol and sim-to-real consistency.}
For each real-world trial we record grasp success, visible failure modes (table collisions, excessive penetration, IK failure), and whether the object remains stably held during lift and small translations (Figs.~\ref{fig:real_run} and~\ref{fig:slanted_obj_collage}).
Object placements are chosen to mirror the MuJoCo scenario grid, enabling direct comparison of success rates and failure modes between simulation and hardware.


\section{Results and Discussion}
\label{sec:results}

\paragraph{Overall performance.}
Across all 81 MuJoCo scenarios, MVB-Grasp more than doubles success on the Z1 arm (59.3\% vs.\ 24.7\%; Table~\ref{tab:z1_dataset_stats}) while using the \emph{same} GraspGen candidates. Per-object gains in Table~\ref{tab:z1_overall} are substantial: cylinders (25.9\,$\rightarrow$\,85.1\%), asymmetric boxes (37.0\,$\rightarrow$\,55.5\%), and bottles (11.1\,$\rightarrow$\,40.0\%). MVBB filtering also reduces the number of candidates forwarded downstream by 10–20\%, improving the quality of the shortlist.

\begin{figure}[ht]
    \centering
    \includegraphics[width=\linewidth]{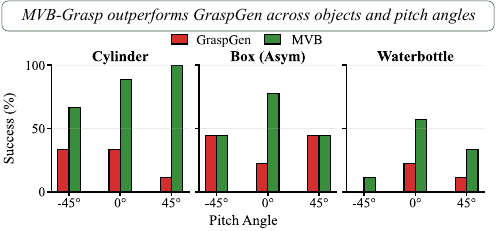}
    \caption{Success rate by object and pitch angle (see Table~\ref{tab:z1_pitch}).}
    \label{fig:z1_pitch_bar}
\end{figure}

\paragraph{Distance and orientation effects.}
Table~\ref{tab:z1_dist_occ} shows that MVB-Grasp consistently improves performance at all distances; for cylinders at \textit{Far} range, success rises from 11.1\% to 88.8\%. Orientation trends in Table~\ref{tab:z1_pitch} and Fig.~\ref{fig:z1_pitch_bar} indicate that MVBB is particularly effective for symmetric objects (cylinder: 11.1\,$\rightarrow$\,100\% at $45^\circ$), gives moderate but configuration-dependent gains for asymmetric boxes, and significantly helps but does not fully solve thin bottle geometry.

\paragraph{Latency and dataset statistics.}
As summarized in Table~\ref{tab:z1_latency_breakdown}, MVBB adds only 6.78\,ms of selection time (2947.4\,ms vs.\ 2940.6\,ms total), confirming that the geometric prior improves reliability with negligible latency overhead. Over 81 episodes per method (Table~\ref{tab:z1_dataset_stats}), this leads to more than twice as many successful trials (48 vs.\ 20), supporting MVB-Grasp as a practical embodiment-aware adapter for existing 6-DoF grasp generators.


\vspace{-0.3cm}
\begin{table}[ht]
    \centering
    \caption{Z1 MuJoCo results aggregated over all positions, occlusions and pitch angles.}
    \label{tab:z1_overall}
    \resizebox{\linewidth}{!}{%
    \begin{tabular}{llrrrr}
        \toprule
        \textbf{Object} & \textbf{Method} & \textbf{\#Ep.} & \textbf{Succ. [\%]} & \textbf{\#Cand.} & \textbf{\#After MVB} \\
        \midrule
        \multirow{2}{*}{Cylinder} & GraspGen & 27 & 25.9 & 35 & -- \\
                                  & \textbf{MVB-Grasp} & 27 & \textbf{85.1} & 35 & 31.0 \\
        \midrule
        \multirow{2}{*}{BoxAsym}  & GraspGen & 27 & 37.0 & 110 & -- \\
                                  & \textbf{MVB-Grasp} & 27 & \textbf{55.5} & 110 & 92.2 \\
        \midrule
        \multirow{2}{*}{Waterbottle} & GraspGen & 27 & 11.1 & 142 & -- \\
                                     & \textbf{MVB-Grasp} & 27 & \textbf{40.0} & 142 & 112.1 \\
        \bottomrule
    \end{tabular}%
    }
\end{table}

\begin{table}[ht]
    \centering
    \caption{Dataset statistics for all logged Z1 MuJoCo episodes.}
    \label{tab:z1_dataset_stats}
    \resizebox{\linewidth}{!}{%
    \begin{tabular}{lrrrr}
        \toprule
        \textbf{Method} & \textbf{\#Ep.} & \textbf{\#Cand. grasps} & \textbf{\#Succ. ep.} & \textbf{\#Fail ep.} \\
        \midrule
        GraspGen & 81 & 7459 & 20 & 61 \\
        \textbf{MVB-Grasp} & 81 & 7459 & \textbf{48} & 33 \\
        \bottomrule
    \end{tabular}%
    }
\end{table}

\begin{table}[ht]
    \centering
    \caption{Latency breakdown of GraspGen and MVB-Grasp over all Z1 MuJoCo scenarios.}
    \label{tab:z1_latency_breakdown}
    \resizebox{\linewidth}{!}{%
    \begin{tabular}{lrrrr}
        \toprule
        \textbf{Method} & \textbf{$t_{\text{gen}}$ [ms]} & \textbf{$t_{\text{coll}}$ [ms]} & \textbf{$t_{\text{sel}}$ [ms]} & \textbf{$t_{\text{total}}$ [ms]} \\
        \midrule
        GraspGen & 463.34 & 2477.28 & 0.02 & 2940.64 \\
        \textbf{MVB-Grasp} & 463.34 & 2477.28 & 6.78 & 2947.40 \\
        \bottomrule
    \end{tabular}%
    }
\end{table}

\begin{table}[ht]
    \centering
    \caption{Effect of distance and occlusion on Z1 MuJoCo grasping performance.}
    \label{tab:z1_dist_occ}
    \resizebox{\linewidth}{!}{%
    \begin{tabular}{lllrrr}
        \toprule
        \textbf{Object} & \textbf{Dist.} & \textbf{Method} & \textbf{Succ. [\%]} & \textbf{\#Cand.} & \textbf{\#After MVB} \\
        \midrule
        \multirow{6}{*}{Cylinder} 
          & Near & GraspGen & 33.3 & 35.0 & -- \\
          & Near & \textbf{MVB-Grasp} & \textbf{88.8} & 35.0 & 34.3 \\
          \cmidrule{2-6}
          & Mid  & GraspGen & 33.3 & 21.1 & -- \\
          & Mid  & \textbf{MVB-Grasp} & \textbf{77.7} & 21.1 & 18.2 \\
          \cmidrule{2-6}
          & Far  & GraspGen & 11.1 & 48.7 & -- \\
          & Far  & \textbf{MVB-Grasp} & \textbf{88.8} & 48.7 & 40.4 \\
        \midrule
        \multirow{6}{*}{BoxAsym} 
          & Near & GraspGen & 55.5 & 75.8 & -- \\
          & Near & \textbf{MVB-Grasp} & 44.4 & 75.8 & 66.6 \\
          \cmidrule{2-6}
          & Mid  & GraspGen & 33.3 & 89.2 & -- \\
          & Mid  & \textbf{MVB-Grasp} & \textbf{77.7} & 89.2 & 79.7 \\
          \cmidrule{2-6}
          & Far  & GraspGen & 22.2 & 163.8 & -- \\
          & Far  & \textbf{MVB-Grasp} & \textbf{44.4} & 163.8 & 130.3 \\
        \midrule
        \multirow{6}{*}{Waterbottle} 
          & Near & GraspGen & 0.0 & 135.3 & -- \\
          & Near & \textbf{MVB-Grasp} & \textbf{14.3} & 135.3 & 104.1 \\
          \cmidrule{2-6}
          & Mid  & GraspGen & 0.0 & 117.0 & -- \\
          & Mid  & \textbf{MVB-Grasp} & \textbf{33.3} & 117.0 & 97.2 \\
          \cmidrule{2-6}
          & Far  & GraspGen & 33.3 & 173.0 & -- \\
          & Far  & \textbf{MVB-Grasp} & \textbf{44.4} & 173.0 & 133.1 \\
        \bottomrule
    \end{tabular}%
    }
\end{table}

\begin{table}[ht]
    \centering
    \caption{Orientation robustness: effect of object pitch for asymmetric objects.}
    \label{tab:z1_pitch}
    \resizebox{\linewidth}{!}{%
    \begin{tabular}{lrlrrr}
        \toprule
        \textbf{Object} & \textbf{Pitch} & \textbf{Method} & \textbf{Succ. [\%]} & \textbf{\#Cand.} & \textbf{\#MVB} \\
        \midrule
        \multirow{6}{*}{Cylinder} 
          & $0^\circ$ & GraspGen & 33.3 & 45.2 & -- \\
          & $0^\circ$ & \textbf{MVB} & \textbf{88.9} & 45.2 & 43.1 \\
          \cmidrule{2-6}
          & $45^\circ$ & GraspGen & 11.1 & 26.0 & -- \\
          & $45^\circ$ & \textbf{MVB} & \textbf{100.0} & 26.0 & 22.4 \\
          \cmidrule{2-6}
          & $-45^\circ$ & GraspGen & 33.3 & 33.5 & -- \\
          & $-45^\circ$ & \textbf{MVB} & \textbf{66.7} & 33.5 & 27.4 \\
        \midrule
        \multirow{6}{*}{BoxAsym} 
          & $0^\circ$ & GraspGen & 22.2 & 115.7 & -- \\
          & $0^\circ$ & \textbf{MVB} & \textbf{77.8} & 115.7 & 92.4 \\
          \cmidrule{2-6}
          & $45^\circ$ & GraspGen & 44.4 & 103.2 & -- \\
          & $45^\circ$ & \textbf{MVB} & 44.4 & 103.2 & 87.8 \\
          \cmidrule{2-6}
          & $-45^\circ$ & GraspGen & 44.4 & 109.8 & -- \\
          & $-45^\circ$ & \textbf{MVB} & 44.4 & 109.7 & 96.2 \\
        \midrule
        \multirow{6}{*}{Bottle} 
          & $0^\circ$ & GraspGen & 22.2 & 126.1 & -- \\
          & $0^\circ$ & \textbf{MVB} & \textbf{57.1} & 126.1 & 109.7 \\
          \cmidrule{2-6}
          & $45^\circ$ & GraspGen & 11.1 & 127.1 & -- \\
          & $45^\circ$ & \textbf{MVB} & \textbf{33.3} & 127.1 & 96.0 \\
          \cmidrule{2-6}
          & $-45^\circ$ & GraspGen & 0.0 & 170.0 & -- \\
          & $-45^\circ$ & \textbf{MVB} & \textbf{11.1} & 170.0 & 130.0 \\
        \bottomrule
    \end{tabular}%
    }
\end{table}

\section{Conclusion}
\label{sec:conclusion}
We studied how to deploy a state-of-the-art 6-DoF grasp generator on a low-cost, workspace-constrained arm in a frontal grasping setting. While GraspGen performs well under standard top-down benchmarks, our Z1 experiments showed many grasps are infeasible for a side-mounted, frontal configuration. The key idea of this paper is to use a simple, embodiment-aware geometric prior, an MVBB/OBB over the segmented object point cloud, to filter and re-score GraspGen candidates without any retraining. The proposed \emph{MVB-Grasp} stack uses MVBB face normals to reject grasps that approach through inaccessible directions and to prioritize candidates aligned with faces reachable from the Z1’s frontal workspace. On an 81-scenario MuJoCo benchmark across three objects, distances, and orientations, MVB-Grasp improves overall success from 24.7\% to 59.3\% (2.4$\times$), with particularly strong gains on symmetric cylinders (25.9\%~$\rightarrow$~85.1\%) and challenging $45^\circ$ orientations (11.1\%~$\rightarrow$~100\%), while adding only 6.78\,ms to the selection stage. MVB-Grasp is not universally better and requires embodiment-specific tuning, but our results show that thin, geometric adapters can substantially improve the practical reliability of powerful but generic grasp generators on constrained, real-world manipulators.

\section*{Acknowledgment}
This work was supported in part by the NYUAD Center for Artificial Intelligence and Robotics (CAIR), funded by Tamkeen under the NYUAD Research Institute Award CG010.

\bibliographystyle{IEEEtran}
\bibliography{cite}

@misc{ma2024physicalpriors,
      title={Generalizing 6-DoF Grasp Detection via Domain Prior Knowledge}, 
      author={Haoxiang Ma and others},
      year={2024},
      eprint={2404.01727},
      archivePrefix={arXiv},
      primaryClass={cs.RO} 
}

@misc{graspgen,
      title={GraspGen: A Diffusion-based Framework for 6-DOF Grasping with On-Generator Training}, 
      author={Adithyavairavan Murali and others},
      year={2025},
      eprint={2507.13097},
      archivePrefix={arXiv},
      primaryClass={cs.RO} 
}

@misc{anygrasp,
      title={AnyGrasp: Robust and Efficient Grasp Perception in Spatial and Temporal Domains}, 
      author={Hao-Shu Fang and others},
      year={2023},
      eprint={2212.08333},
      archivePrefix={arXiv},
      primaryClass={cs.RO} 
}

@misc{contactgraspnet,
      title={Contact-GraspNet: Efficient 6-DoF Grasp Generation in Cluttered Scenes}, 
      author={Martin Sundermeyer and others},
      year={2021},
      eprint={2103.14127},
      archivePrefix={arXiv},
      primaryClass={cs.RO} 
}

@misc{mahler2017gpd,
      title={Grasp Pose Detection in Point Clouds}, 
      author={Andreas ten Pas and others},
      year={2017},
      eprint={1706.09911},
      archivePrefix={arXiv},
      primaryClass={cs.RO} 
}

@misc{newbury2022review,
      title={Deep Learning Approaches to Grasp Synthesis: A Review}, 
      author={Rhys Newbury and others},
      year={2023},
      eprint={2207.02556},
      archivePrefix={arXiv},
      primaryClass={cs.RO} 
}

@misc{heatmap6dof,
      title={Efficient Heatmap-Guided 6-Dof Grasp Detection in Cluttered Scenes}, 
      author={Siang Chen and others},
      year={2024},
      eprint={2403.18546},
      archivePrefix={arXiv},
      primaryClass={cs.RO} 
}

@misc{Mousavian6DoFGraspNetICCV2019,
      title={6-DOF GraspNet: Variational Grasp Generation for Object Manipulation}, 
      author={Arsalan Mousavian and others},
      year={2019},
      eprint={1905.10520},
      archivePrefix={arXiv},
      primaryClass={cs.CV} 
}

@misc{LenzGraspIJRR2015,
      title={Deep Learning for Detecting Robotic Grasps}, 
      author={Ian Lenz and others},
      year={2014},
      eprint={1301.3592},
      archivePrefix={arXiv},
      primaryClass={cs.LG} 
}

@misc{MorrisonGGCNN2018,
      title={Closing the Loop for Robotic Grasping: A Real-time, Generative Grasp Synthesis Approach}, 
      author={Douglas Morrison and others},
      year={2018},
      eprint={1804.05172},
      archivePrefix={arXiv},
      primaryClass={cs.RO} 
}

@misc{MahlerDexNet2RSS2017,
      title={Dex-Net 2.0: Deep Learning to Plan Robust Grasps with Synthetic Point Clouds and Analytic Grasp Metrics}, 
      author={Jeffrey Mahler and others},
      year={2017},
      eprint={1703.09312},
      archivePrefix={arXiv},
      primaryClass={cs.RO} 
}

@article{LiOBBLightGBM2021,
      author={Lin, Shifeng and others},
      year = {2023},
      month = {06},
      pages = {1-17},
      title = {Robot grasping based on object shape approximation and LightGBM},
      volume = {83},
      journal = {Multimedia Tools and Applications},
      doi = {10.1007/s11042-023-15547-y}
}

@incollection{GeidenstamBoxBasedRSS2009,
    author={Geidenstam, Sebastian and others},
    isbn = {9780262289801},
    title = {Learning of 2D Grasping Strategies from Box-based 3D Object Approximations},
    booktitle = {Robotics: Science and Systems V},
    publisher = {The MIT Press},
    year = {2010},
    month = {07},
    doi = {10.7551/mitpress/8727.003.0003},
    eprint = {https://direct.mit.edu/book/chapter-pdf/2277302/9780262289801_cab.pdf}
}

@article{VoOBBYOLO2025,
      title = {Improving robotic grasping accuracy through oriented bounding box detection with YOLOv11-OBB},
      journal = {Heliyon},
      volume = {11},
      number = {12},
      pages = {e43512},
      year = {2025},
      issn = {2405-8440},
      doi = {https://doi.org/10.1016/j.heliyon.2025.e43512},
      author={Vo Duy Cong and others}
}

@misc{ZhangGoalGrasp2024,
      title={GoalGrasp: Grasping Goals in Partially Occluded Scenarios without Grasp Training}, 
      author={Shun Gui and others},
      year={2025},
      eprint={2405.04783},
      archivePrefix={arXiv},
      primaryClass={cs.RO} 
}

@misc{RedmonGraspICRA2015,
      title={Real-Time Grasp Detection Using Convolutional Neural Networks}, 
      author={Joseph Redmon and others},
      year={2015},
      eprint={1412.3128},
      archivePrefix={arXiv},
      primaryClass={cs.RO} 
}

@INPROCEEDINGS{HuebnerMVBBICRA2008,
  author={Huebner, Kai and others},
  booktitle={2008 IEEE International Conference on Robotics and Automation}, 
  title={Minimum volume bounding box decomposition for shape approximation in robot grasping}, 
  year={2008},
  volume={},
  number={},
  pages={1628-1633},
  doi={10.1109/ROBOT.2008.4543434}
}

@software{yolov8_ultralytics,
  author={Glenn Jocher and others},
  title = {Ultralytics YOLOv8},
  version = {8.0.0},
  year = {2023},
  orcid = {0000-0001-5950-6979, 0000-0002-7603-6750, 0000-0003-3783-7069},
  license = {AGPL-3.0}
}

\end{document}